\documentclass[letterpaper]{article}

\usepackage[preprint]{aaai2027}
\usepackage[hyphens]{url}
\usepackage{graphicx}
\graphicspath{{./figures/}}
\usepackage{natbib}
\usepackage{caption}
\usepackage{booktabs}

\usepackage{amsmath}
\usepackage{amssymb}
\usepackage{bm}
\usepackage{mathtools}

\usepackage{array}
\usepackage{tabularx}
\usepackage{multirow}
\usepackage{makecell}
\usepackage{ifpdf}
\usepackage[table]{xcolor}
\definecolor{resultgray}{gray}{0.92}

\newcommand{\passatk}[1]{pass@#1}

\begin{document}
\title{TaPR: Test-Aware Policy Refinement for Feedback-Conditioned Code Generation}

\ifpdf
  % PDF: preserve the native AAAI author-note formatting.
  \author{
    Aofan Liu\equalcontrib,
    Jingxiang Meng\equalcontrib,
    Fangxin Liu\corresponding,
    Yongbiao Chen\corresponding
  }
  \affiliations{}
\else
  % arXiv HTML: use simple visible markers.
  \author{Aofan Liu\textsuperscript{*}, Jingxiang Meng\textsuperscript{*}, Fangxin Liu\textsuperscript{\textdagger}, Yongbiao Chen\textsuperscript{\textdagger}}
  \date{}
\fi

\maketitle

\ifpdf\else
  \begingroup
  \centering
  \footnotesize
  \textsuperscript{*} These authors contributed equally.
  \textsuperscript{\textdagger} Corresponding authors.
  \par
  \endgroup
\fi

\begin{abstract}
Multi-turn code agents rely on execution feedback to repair incorrect programs, yet standard reinforcement learning paradigms optimize and evaluate policy performance primarily using single-shot outcome rewards. This misalignment conflates initial code generation with feedback-driven refinement, discards granular execution signals across intermediate turns, and fails to evaluate whether the policy actually acquires self-repair capabilities. We propose Test-aware Policy Refinement (TaPR), a framework that transforms execution feedback into a dense per-turn test-pass-ratio reward under a consistent multi-turn interaction protocol. Across six models on 219 code-generation problems from LiveCodeBench, TaPR improves the pooled three-turn success rate ($\text{Pass}@3$) by $2.44$ percentage points. In the predefined 7B/8B high-headroom slice, pooled accuracy increases from 30.25\% to 33.56\% (+3.31 pp), with $42$ improvements and $13$ regressions in paired trials. On a matched Qwen3-8B ablation, the dense reward supplies nonzero feedback in all of the first ten steps and reaches a higher Hard-subset peak than outcome-only GRPO within the tested budget, although GRPO nearly matches pooled $\text{Pass}@3$ by step 300. Our primary contribution is a reward-decomposition framework and a turn-aware evaluation protocol that decouple first-shot generation quality from multi-turn repair competence.
\end{abstract}

\section{Introduction}
\label{sec:intro}

In reinforcement learning (RL) for mathematical and logical reasoning, process supervision often outperforms outcome supervision by assigning explicit credit to intermediate reasoning steps \cite{lightman2023verify,wang2024mathshepherd}. Code generation presents a uniquely suitable environment for process-level feedback: each interactive turn yields an \emph{executable} program whose functional correctness can be evaluated by a deterministic test oracle. Nevertheless, contemporary code RL paradigms remain predominantly outcome-driven. Both their optimization objectives and evaluation metrics reduce multi-turn interactions to final-answer correctness. Consequently, standard metrics such as \passatk{k} indicate overall task success without identifying whether the model actively repaired its code using execution feedback. This framework creates a fundamental \emph{optimization--evaluation mismatch}, where feedback-conditioned refinement is neither explicitly rewarded nor cleanly evaluated as a distinct capability.

A standard multi-turn code interaction operates as a four-step loop: the policy generates a candidate program, executes it, receives test feedback, and revises the code. After each turn, the execution oracle produces a fine-grained signal representing the proportion of passed unit tests. Outcome-based code RL discards these intermediate execution signals, rewarding only ultimate correctness at the final turn. As a result, standard formulations fail to reward incremental improvements across successive turns, obscuring whether a policy is developing genuine self-repair competence or merely relying on first-shot luck.

In this work, we investigate whether first-shot generation quality and feedback-conditioned refinement can be disentangled, trained, and evaluated as distinct targets. While final-outcome objectives optimize end-to-end success, they offer no direct gradient credit for intermediate progress. To address this limitation, we introduce \textbf{Test-aware Policy Refinement (TaPR)}, a framework that converts intermediate hidden-test pass ratios into dense, per-turn rewards under a unified multi-turn interaction protocol. 

Through controlled experiments on Qwen3-8B under matched data, optimizer, batch size, and KL constraints, we observe a notable empirical phenomenon. The untrained baseline, standard GRPO, and TaPR yield identical first-turn accuracy ($\text{Pass}@1 = 7/61$) on the Hard problem subset, but their three-turn performance ($\text{Pass}@3$) differs at $7/61$, $8/61$, and $10/61$, respectively. The observed difference appears under feedback-conditioned evaluation rather than single-shot generation, a pattern we term the \textbf{$\text{Pass}@1/\text{Pass}@3$ dissociation}.

Extensive evaluations across six open-weight language models demonstrate that TaPR provides consistent, positive performance gains. In the 7B/8B high-headroom regime (comprising Qwen2.5-7B, Qwen2.5-7B-Instruct, Llama-3.1-8B, and Qwen3-8B), TaPR increases the pooled LCB-219 $\text{Pass}@3$ metric by $+3.31$ percentage points ($+10.9\%$ relative gain), winning $42$ head-to-head problem comparisons against $13$ losses compared to the baseline. We employ this high-headroom regime to highlight specific optimization dynamics rather than as a selective filtering heuristic.

While prior work such as SCoRe \cite{kumar2024score} demonstrated that supervised fine-tuning (SFT) for self-correction can cause behavior collapse where later turns fail to improve over initial attempts, our findings address a distinct challenge. Under a fully matched multi-turn RL setting, we demonstrate that per-turn dense rewards and sparse outcome rewards induce fundamentally different optimization trajectories and performance profiles. This aligns with recent insights showing that evaluation metrics in single-turn losses actively shape the underlying capabilities being optimized \cite{chen2025passatk}; our work extends this principle to multi-turn code RL by establishing the critical role of reward decomposition.

\begin{figure*}[t]
\centering
\includegraphics[width=\textwidth]{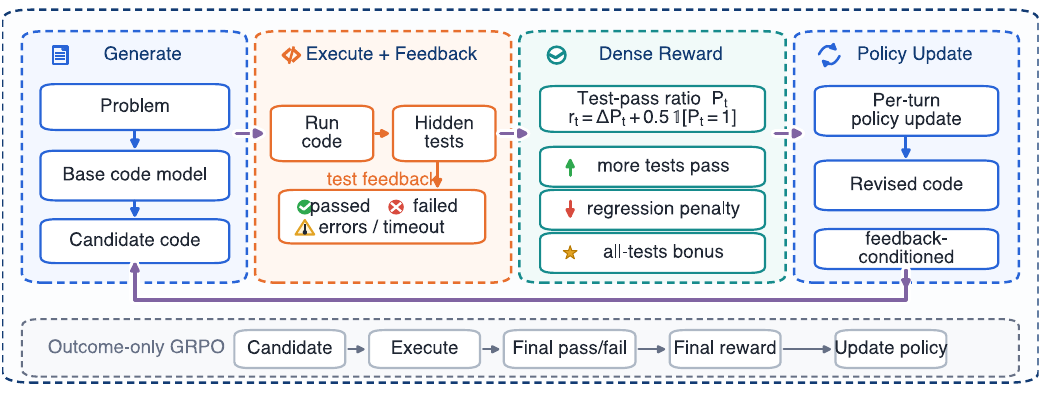}
\caption{Overview of Test-aware Policy Refinement (TaPR). The upper path illustrates multi-turn code generation where execution feedback is converted into a dense per-turn test-pass reward; the lower path contrasts standard GRPO, which provides only sparse final-answer feedback.}
\label{fig:architecture}
\end{figure*}

\paragraph{Observations.}
To verify that the observed dissociation stems from the underlying reward formulation rather than transient checkpoint artifacts, we perform three orthogonal diagnostic analyses:
\begin{itemize}
  \item \emph{Training Curve Dynamics}: On Qwen3-8B, TaPR reaches its peak $\text{Pass}@3$ score at step 300 ($53.4\%$, representing a $+1.83$ percentage-point overall increase and a Hard-subset improvement from $7$ to $10$) before returning to baseline by step 500. The earlier rise is consistent with improved sample efficiency within the tested budget.
  \item \emph{Reward Signal Properties}: Under identical hyperparameter configurations, the reward variance and magnitude under GRPO exhibit a $13\times$ lower mean and standard deviation, accompanied by a $10\times$ lower early-training gradient norm relative to the dense per-turn reward of TaPR.
  \item \emph{Cross-Model Generalization}: The selected TaPR checkpoints are positive for all six evaluated models, with the 7B/8B cohort improving pooled $\text{Pass}@3$ accuracy from $30.25\%$ to $33.56\%$.
\end{itemize}

\paragraph{Contributions.}
Our contributions are summarized as follows:
\begin{itemize}
  \item \textbf{Empirical Discovery of Dissociation.} We identify and formalize the $\text{Pass}@1/\text{Pass}@3$ dissociation phenomenon, showing that under strictly matched training conditions, dense per-turn rewards improve iterative repair capability ($\text{Pass}@3$) without altering initial generation accuracy ($\text{Pass}@1$).
  \item \textbf{Dense Reward Formulation (TaPR).} We propose Test-aware Policy Refinement, a practical multi-turn RL approach that transforms intermediate execution pass ratios into dense rewards, resolving credit assignment ambiguity in code refinement.
  \item \textbf{Mechanism and Efficiency Insights.} Training curves and reward-gradient statistics show that dense execution rewards provide a less sparse early optimization signal and reach a higher Hard-subset peak within the tested budget.
  \item \textbf{Broad Benchmark Validation.} We evaluate TaPR across six language models from three model families (7B--14B), reporting both the positive pooled effect and the limited changes observed for the two 14B models.
\end{itemize}

\section{Related Work}
\label{sec:related}

\paragraph{Execution-Based Benchmarks and Code Evaluation.}
Modern code generation benchmarks, including HumanEval, MBPP, APPS, BigCodeBench, and LiveCodeBench, evaluate functional correctness by executing generated candidate programs against test suites \cite{chen2021humaneval,austin2021mbpp,hendrycks2021apps,zhuo2024bigcodebench,jain2024livecodebench}. While conventional pipelines restrict execution feedback exclusively to final evaluation, TaPR directly integrates intermediate hidden-test pass ratios into the training loop as dense, per-turn reinforcement signals.

\paragraph{Execution-Feedback Training for Code.}
Self-Edit trains a fault-aware editor to revise programs from execution results, while LETI iteratively fine-tunes a generator on textual interpreter feedback \cite{zhang2023selfedit,wang2024leti}. CodePRM trains an execution-aware process verifier for generate--verify--refine search \cite{li2025codeprm}. Among policy-training approaches, CodeRL employs an actor-critic architecture with a learned critic \cite{le2022coderl}, RLEF uses PPO for iterative code generation \cite{gehring2025rlef}, and $\mu$Code uses iterative imitation learning with single-step rewards and a learned verifier \cite{jain2025mucode}. CoCoS combines an accumulated trajectory objective with fine-grained correction rewards for small code models \cite{cho2025cocos}. These methods establish that training can improve feedback use; TaPR isolates a narrower question by comparing dense per-turn and sparse outcome rewards under matched architecture, data, optimizer, and KL constraints.

\paragraph{Self-Correction and Behavior Collapse.}
In SFT-based self-correction, SCoRe \cite{kumar2024score} revealed a critical behavior collapse phenomenon wherein trained policies improve initial generation quality while leaving subsequent refinement turns unoptimized. To address this issue, multi-turn RL was introduced to sustain correction gains. Our study offers a precise, complementary insight within a matched multi-turn RL comparison: dense per-turn rewards and sparse outcome rewards produce distinct performance trajectories. On the Qwen3-8B Hard subset, TaPR reaches a higher multi-turn refinement score ($\text{Pass}@3$) within the tested budget while single-shot accuracy ($\text{Pass}@1$) remains unchanged.

\paragraph{Concurrent Multi-Turn GRPO Systems.}
MURPHY and ReVeal extend GRPO-style optimization with feedback-conditioned rollout trees, retrospective credit assignment, or self-verification \cite{ekbote2025murphy,jin2025reveal}. MM-ReCoder applies a two-stage self-correction strategy to chart-to-code generation \cite{tang2026mmrecoder}, a different task from text-only competitive programming. TaPR focuses on reward decomposition rather than a new verifier or rollout topology. Its turn-aware $\text{Pass}@1/\text{Pass}@3$ analysis measures whether training changes first-shot generation or feedback-conditioned repair.

\paragraph{Dense Rewards and Process Supervision.}
Dense reward design does not universally guarantee performance improvements. Studies such as Iterative Reward Calibration demonstrate that dense rewards can misalign policy advantage directions in general agent tasks \cite{modecrua2026irc}, while WebAgent-R1 observes scenarios where sparse rewards outperform dense signals in web navigation \cite{wei2025webagentr1}. TaPR mitigates potential misalignment by sourcing its dense reward from deterministic, external test-suite execution rather than heuristic reward shaping. Furthermore, unlike Process Reward Models (PRMs) in mathematical reasoning that depend on learned verifiers \cite{lightman2023verify,wang2024mathshepherd}, TaPR leverages the deterministic nature of code execution, using the test oracle directly to furnish non-parametric intermediate rewards.

\paragraph{Prompted Refinement and Capacity Amplification.}
Inference-time refinement frameworks like Self-Refine and Reflexion rely on fixed model weights, using structured prompting to elicit self-correction \cite{madaan2023selfrefine,shinn2023reflexion}. However, recent surveys show that ungrounded self-correction without external execution feedback remains unreliable \cite{kamoi2024selfcorrection}. TaPR embeds refinement directly into policy weights through RL. This distinction relates to broader investigations into whether RL creates new capabilities or amplifies capabilities already present in base models \cite{yue2025rlvr}. Our larger gains on models with partially correct outputs are consistent with amplification, but the present experiments do not distinguish that explanation from scale-specific optimization or data-distribution effects.

\section{Methodology}
\label{sec:method}

We present Test-aware Policy Refinement (TaPR), a multi-turn reinforcement learning framework designed to resolve credit assignment ambiguity in feedback-driven code generation. Figure~\ref{fig:architecture} provides an architectural comparison between TaPR and standard GRPO. While outcome-based methods condense multi-turn interactions into a sparse final reward, TaPR transforms intermediate hidden-test execution signals into dense per-turn learning objectives.

\subsection{Multi-Turn Interaction Formulation}
\label{subsec:multi-turn-rollout}

We model multi-turn code repair as a Markov Decision Process (MDP). Given a programming problem specification $x$ and an associated execution oracle with hidden tests, the policy $\pi_\theta$ interacts with the environment for up to $T$ turns. At turn $t \in \{1, \ldots, T\}$, the context $c_t = (x, a_1, e_1, \ldots, a_{t-1}, e_{t-1})$ incorporates the original problem statement, historical candidate code attempts $\{a_i\}_{i=1}^{t-1}$, and execution feedback $\{e_i\}_{i=1}^{t-1}$ returned by the test sandbox. The policy then samples a revised code attempt $a_t \sim \pi_\theta(\cdot \mid c_t)$, which is subsequently evaluated by the test runner to generate updated execution feedback $e_t$. The feedback $e_t$ comprises standard output/error logs and pass/fail execution indicators across all hidden unit tests.

\subsection{Dense Per-Turn Reward Formulation}
\label{subsec:dense-reward}

Let $m$ denote the total number of hidden test cases associated with problem $x$. Upon executing candidate solution $a_t$ at turn $t$, the execution oracle yields a binary test-pass vector $\mathbf{s}_t = (s_{t,1}, s_{t,2}, \ldots, s_{t,m})^\top \in \{0,1\}^m$, where $s_{t,j} = 1$ if test $j$ passes and $0$ otherwise. Semantic errors, incorrect output assertions, compilation failures, and runtime timeouts explicitly result in $s_{t,j} = 0$.

We define the primary per-turn execution quality signal $P_t$ as the proportion of passed hidden tests:
\begin{equation}
P_t = \frac{1}{m} \sum_{j=1}^{m} s_{t,j}, \quad \text{with } P_0 = 0.
\label{eq:pass-ratio}
\end{equation}
More generally, $P_t$ can incorporate an infrastructure validity mask $\mathbf{m}_t \in \{0,1\}^m$ and test importance weights $W = (W_1, \ldots, W_m)^\top$ via $P_t = \frac{W^\top (\mathbf{m}_t \odot \mathbf{s}_t)}{W^\top \mathbf{m}_t}$, though we set $W_j = 1$ and $m_{t,j} = 1$ across all experiments to maintain unweighted pass ratios.

To reward incremental functional progress across successive interaction turns, we compute the turn-level pass ratio improvement $\Delta P_t$:
\begin{equation}
\Delta P_t = P_t - P_{t-1}.
\label{eq:delta-pass}
\end{equation}

The complete TaPR reward $r_t$ assigned to action $a_t$ at turn $t$ is formulated as a two-term decomposition combining progress-based reward shaping and a task completion bonus:
\begin{equation}
r_t = \lambda_{\Delta} \Delta P_t + \lambda_{\mathrm{done}} \mathbb{1}[P_t = 1],
\label{eq:tapr-reward}
\end{equation}
where $\mathbb{1}[\cdot]$ is the indicator function, $\lambda_{\Delta} = 1.0$ controls the sensitivity to pass-ratio updates, and $\lambda_{\mathrm{done}} = 0.5$ provides an explicit incentive for full functional correctness.

Under this formulation, the policy receives explicit gradient feedback whenever code revision alters the passed test fraction, rather than exclusively when full correctness is achieved. In our empirical setups, TaPR maintains a well-conditioned mean step reward between $0.10$ and $0.17$ with a standard deviation of $0.27$--$0.37$. In contrast, GRPO under identical settings yields a sparse mean step reward of $0.018$, with over $33\%$ of optimization steps receiving zero reward signal.

\paragraph{Reward semantics.}
The difference term rewards newly passed tests and penalizes regressions. For an undiscounted trajectory, it telescopes:
\begin{equation}
\sum_{t=1}^{T}\Delta P_t=P_T-P_0=P_T.
\label{eq:telescoping}
\end{equation}
Temporary progress that is later lost therefore does not inflate return. The completion bonus is issued once because solved trajectories terminate. Unlike an absolute per-turn reward $P_t$, which repeatedly credits already-passed tests, $\Delta P_t$ preserves final functional quality while locating credit on the revision that changes it.

\section{Experiments}

\begin{table*}[t]
\centering
{\small
\setlength{\tabcolsep}{3pt}
\begin{tabularx}{\textwidth}{llXcc}
\toprule
Model & Method & Benchmark / metric / protocol & Result & $\Delta$pp \\
\midrule
\emph{Qwen2.5-7B} & \emph{Baseline (ours)} & \emph{LCB-219 \passatk{3}, 3 turns} & $\mathit{24.20}$ & $0.00$ \\
\rowcolor{resultgray}
\textbf{Qwen2.5-7B} & \textbf{TaPR} & \textbf{LCB-219 \passatk{3}, 3 turns} & $\mathbf{30.14}$ & $\mathbf{+5.94}$ \\
\midrule
\emph{Qwen2.5-7B-Instruct} & \emph{Baseline \cite{lee2026mapcoderlite}} & \emph{APPS-150 \passatk{1}, greedy} & $\mathit{4.00}$ & $0.00$ \\
Qwen2.5-7B-Instruct & MapCoder \cite{lee2026mapcoderlite} & APPS-150 $\text{Pass}@1$, multi-agent & $6.00$ & $+2.00$ \\
Qwen2.5-7B-Instruct & MapCoder-Lite \cite{lee2026mapcoderlite} & APPS-150 $\text{Pass}@1$, agent-wise LoRA & $8.00$ & $+4.00$ \\
\emph{Qwen2.5-7B} & \emph{TaPR start (ours)} & \emph{APPS-300 \passatk{2}, 2 turns} & $\mathit{18.33}$ & $+14.33$ \\
\rowcolor{resultgray}
\textbf{Qwen2.5-7B} & \textbf{TaPR} & \textbf{APPS-300 \passatk{2}, 2 turns} & $\mathbf{21.00}$ & $\mathbf{+17.00}$ \\
\midrule
\emph{Qwen2.5-7B-Instruct} & \emph{Baseline \cite{hu2025quest}} & \emph{LCB-v5 \passatk{1}} & $\mathit{14.3}$ & $0.0$ \\
Qwen2.5-7B-Instruct & TACO RL \cite{hu2025quest} & LCB-v5 $\text{Pass}@1$ & $17.3$ & $+3.0$ \\
Qwen2.5-7B-Instruct & QueST \cite{hu2025quest} & LCB-v5 $\text{Pass}@1$ & $18.6$ & $+4.3$ \\
\rowcolor{resultgray}
\textbf{Qwen2.5-7B-Instruct} & \textbf{TaPR} & \textbf{LCB-219 \passatk{3}, 3 turns} & $\mathbf{27.85}$ & $\mathbf{+13.55}$ \\
\midrule
\emph{Llama-3.1-8B-Instruct} & \emph{Baseline \cite{lin2026macscoder}} & \emph{LCB-v5 \passatk{1}} & $\mathit{10.7}$ & $0.0$ \\
Llama-3.1-8B & SOL-VER \cite{lin2025solver} & LiveCodeBench $\text{Pass}@1$, greedy & $27.24$ & $+16.54$ \\
Llama-3.1-8B-Instruct & MACS-Coder \cite{lin2026macscoder} & LCB-v5 $\text{Pass}@1$, 5 planning iters & $15.1$ & $+4.4$ \\
\rowcolor{resultgray}
\textbf{Llama-3.1-8B} & \textbf{TaPR} & \textbf{LCB-219 \passatk{3}, 3 turns} & $\mathbf{22.83}$ & $\mathbf{+12.13}$ \\
\midrule
\emph{Qwen3-8B} & \emph{Baseline \cite{chen2026cobalt}} & \emph{LCB Jan--Apr.\ 2025 \passatk{1}} & $\mathit{26.8}$ & $0.0$ \\
Qwen3-8B & GRPO-MT \cite{chen2026cobalt} & LCB Jan--Apr.\ 2025 $\text{Pass}@1$, $t0{\to}t3$ & $30.5$ & $+3.7$ \\
Qwen3-8B & VeRPO-MT \cite{chen2026cobalt} & LCB Jan--Apr.\ 2025 $\text{Pass}@1$, $t0{\to}t3$ & $33.1$ & $+6.3$ \\
Qwen3-8B & COBALT \cite{chen2026cobalt} & LCB Jan--Apr.\ 2025 $\text{Pass}@1$, $t0{\to}t3$ & $34.9{\pm}0.3$ & $+8.1$ \\
Qwen3-8B & X-Coder SFT \cite{wu2026xcoder} & LCB-v6 $\text{Avg}@8$, synthetic SFT & $55.4{\pm}2.3$ & $+28.6$ \\
Qwen3-8B & X-Coder \cite{wu2026xcoder} & LCB-v6 $\text{Avg}@8$, SFT$\to$RL & $56.5{\pm}1.3$ & $+29.7$ \\
Qwen3-8B & MACS-Coder \cite{lin2026macscoder} & LCB-v5 $\text{Pass}@1$, 5 planning iters & $63.1$ & $+36.3$ \\
\rowcolor{resultgray}
\textbf{Qwen3-8B} & \textbf{TaPR} & \textbf{LCB-219 \passatk{3}, 3 turns} & $\mathbf{53.42}$ & $\mathbf{+26.62}$ \\
\midrule
\emph{Qwen3-14B} & \emph{Baseline \cite{lin2026macscoder}} & \emph{LCB-v5 \passatk{1}} & $\mathit{43.5}$ & $0.0$ \\
Qwen3-14B & MACS-Coder \cite{lin2026macscoder} & LCB-v5 $\text{Pass}@1$, 5 planning iters & $68.6$ & $+25.1$ \\
\rowcolor{resultgray}
\textbf{Qwen3-14B} & \textbf{TaPR} & \textbf{LCB-219 \passatk{3}, 3 turns} & $\mathbf{59.36}$ & $\mathbf{+15.86}$ \\
\midrule
\emph{Qwen2.5-Coder-14B} & \emph{Baseline (ours)} & \emph{LCB-219 \passatk{3}, 3 turns} & $\mathit{41.55}$ & $0.00$ \\
\rowcolor{resultgray}
\textbf{Qwen2.5-Coder-14B} & \textbf{TaPR} & \textbf{LCB-219 \passatk{3}, 3 turns} & $\mathbf{42.01}$ & $\mathbf{+0.46}$ \\
\bottomrule
\end{tabularx}
}
\caption{Public results grouped by backbone. Within each model--benchmark-family block, $\Delta$pp is measured from the lowest displayed untrained checkpoint. Cross-protocol differences are descriptive because benchmarks and inference protocols differ. TaPR uses LCB-219 ($n=219$ per model), except for APPS-300.}
\label{tab:matrix}
\end{table*}

\label{sec:experiments}

To evaluate whether TaPR successfully decouples first-shot generation from multi-turn code repair, we conduct extensive experiments across diverse model families and parameter scales.

\subsection{Experimental Setup}
\label{subsec:exp-setup}

\paragraph{Datasets and Evaluation Protocol.}
The training dataset comprises $500$ standard input/output (stdin-style) algorithmic programming problems selected from the APPS benchmark \cite{hendrycks2021apps}. For each problem instance, multi-turn interaction trajectories up to $T=3$ turns are sampled using the target base model, yielding a curated pool of approximately $1300$ to $1400$ training rollouts; trajectories that achieve full test compliance at turn $t$ trigger early termination to avoid redundant iterations. For out-of-distribution evaluation, we utilize the stdin-only subset of LiveCodeBench \cite{jain2024livecodebench}, designated as LCB-219 under a standardized three-turn interactive evaluation protocol. We additionally report an in-domain held-out check on $300$ APPS problems for Qwen2.5-7B.

\paragraph{Training Implementation and Hyperparameters.}
We implement our training framework using the \texttt{GRPOTrainer} module within TRL, following the optimization design of DeepSeekMath \cite{shao2024deepseekmath}. Parameter-efficient fine-tuning is conducted using LoRA adapters ($r=32$, $\alpha=64$) \cite{hu2021lora} on a cluster node equipped with four NVIDIA A100 GPUs. The primary training configuration uses a peak learning rate of $5 \times 10^{-6}$ governed by a cosine decay schedule, a KL-divergence penalty weight of $\beta = 0.04$, a per-device batch size of $4$ with $4$ gradient accumulation steps, and $4$ completion rollouts per prompt up to a maximum generation length of $1024$ tokens per turn, optimized across $500$ total steps. An exploratory 14B stress setting ($LR = 5 \times 10^{-5}, \beta = 0.005$) regressed by $9.6$\,pp relative to its untrained checkpoint. We therefore use the conservative configuration throughout the reported comparison; the stress run does not establish which hyperparameter caused the regression.

\paragraph{Checkpoint reporting and paired analysis.}
For the six-model result, each TaPR checkpoint is paired with its own untrained initialization on the same $219$ problem identifiers. Predictions are aligned by \texttt{question\_id} before pooling model--problem pairs. We report the highest observed $\text{Pass}@3$ checkpoint among saved checkpoints, so these are best-observed rather than independently selected estimates. The supplementary document exposes the available trajectories; paired tests quantify problem-level uncertainty but not checkpoint-selection uncertainty.

\subsection{Cross-Model Performance}
\label{subsec:main-results}

Table~\ref{tab:matrix} groups results by model family and uses the lowest displayed untrained checkpoint in each model--benchmark-family block as its Baseline. Every $\Delta$pp entry is recomputed from that row, including TaPR and public methods. This normalization makes the numerical reference explicit, but it does not make heterogeneous protocols equivalent: LCB releases, APPS splits, $\text{Pass}@k$, prompts, and inference budgets still differ. We therefore use the public rows to locate TaPR within the reported range for the same backbone family, not to claim a controlled ranking. Controlled evidence comes from our six paired LCB-219 evaluations: TaPR improves pooled $\text{Pass}@3$ by $+2.44$\,pp, with $51$ baseline-fail/TaPR-pass cases against $19$ regressions across $1{,}314$ model--problem pairs ($p=8.30\times10^{-5}$, one-sided exact test).

We found no public result with the same Qwen2.5-7B base checkpoint, APPS-300 split, and two-turn $\text{Pass}@2$ protocol. The closest APPS-family comparison uses Qwen2.5-7B-Instruct on a random 150-problem split with greedy $\text{Pass}@1$ \cite{lee2026mapcoderlite}, so Table~\ref{tab:matrix} reports it as context rather than a head-to-head baseline. On our APPS-300 held-out set, Qwen2.5-7B $\text{Pass}@1$ increases from $49/300$ ($16.33\%$) to $56/300$ ($18.67\%$), $\text{Pass}@2$ from $55/300$ ($18.33\%$) to $63/300$ ($21.00\%$), and $\text{Pass}@3$ from $58/300$ ($19.33\%$) to $63/300$ ($21.00\%$). The retained APPS evaluation files contain aggregate counts but not per-problem paired outcomes, so we report this result as a secondary check and exclude it from McNemar and bootstrap analyses.

The paired TaPR results show their clearest concentration in the 7B/8B regime (Qwen2.5-7B, Qwen2.5-7B-Instruct, Llama-3.1-8B, and Qwen3-8B). Within this predefined scale slice, pooled $\text{Pass}@3$ rises from $30.25\%$ to $33.56\%$ ($+3.31$\,pp; $+10.9\%$ relative), with $42$ improvements and $13$ regressions across $876$ paired trials. The two Qwen2.5-7B variants improve by $+4.79$\,pp in aggregate. By contrast, the paired changes for Qwen3-14B and Qwen2.5-Coder-14B are $+0.91$ and $+0.46$\,pp, respectively; these small changes indicate limited observed headroom, not a demonstrated ceiling effect. Per-model intervals, Fisher's meta-analysis, and difficulty-stratified counts appear in Supplementary Section~3.

Table~\ref{tab:matrix} uses one arithmetic convention for visual comparison: within each model--benchmark-family block, every displayed $\Delta$pp is measured from the lowest displayed untrained checkpoint. Except where the table explicitly shows our matched LCB-219 baseline, a TaPR $\Delta$pp in this table is therefore \emph{not} the controlled gain from TaPR training. It is a distance from the block's public reference value, sometimes under a different LCB release or interaction protocol. The controlled TaPR effect is the separately reported before--after change against each model's own untrained checkpoint on the same $219$ problems. This distinction prevents the literature rows from being interpreted as a common leaderboard.

\subsection{Training Trajectory and Overfitting Dynamics}
\label{sec:qwen38b-curve}

\begin{figure*}[t]
\centering
\includegraphics[width=0.75\textwidth]{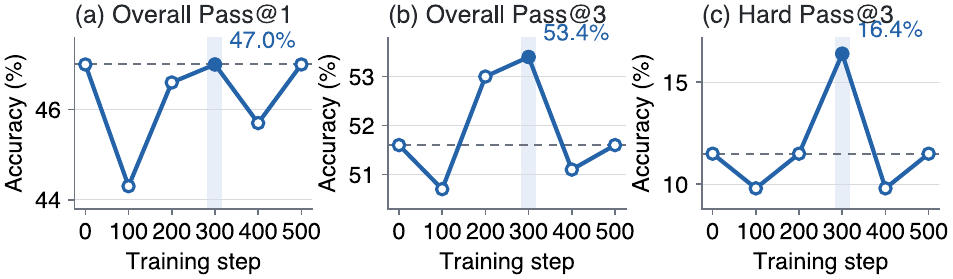}
\caption{Qwen3-8B \passatk{1} and \passatk{3} on LCB-219 across checkpoints. Dashed lines denote the untrained checkpoint and shading marks the selected step 300. The \passatk{3} curve peaks at step 300 and returns to baseline at step 500; \passatk{1} is identical at steps 0, 300, and 500.}
\label{fig:training-curve}
\end{figure*}

Figure~\ref{fig:training-curve} tracks Qwen3-8B performance across 500 training steps on LCB-219. On the overall benchmark, $\text{Pass}@3$ peaks at step 300 ($53.4\%$, a $+1.8$\,pp gain over baseline) before returning to baseline performance by step 500. This non-monotonic trajectory is driven primarily by the Hard problem subset, where $\text{Pass}@3$ improves from $7/61$ at baseline to $10/61$ at step 300, subsequently reverting to $7/61$ at step 500. Crucially, single-shot accuracy ($\text{Pass}@1$) remains flat at $47.0\%$ across steps 0, 300, and 500, confirming that the transient gain stems exclusively from feedback-conditioned refinement rather than initial generation quality.

\begin{table*}[t]
\centering
{\small
\begin{tabular}{llccc}
\toprule
Stage & Measurement & TaPR & Outcome GRPO & Difference \\
\midrule
Availability & zero-reward steps & $0/10$ & $4/10$ & fewer null updates \\
Signal scale & step-1 reward mean & $0.160$ & $0.0125$ & $13\times$ larger \\
Signal scale & step-1 reward std & $0.320$ & $0.025$ & $13\times$ larger \\
Optimization & early gradient norm & $0.030$ & $0.003$ & $10\times$ larger \\
\bottomrule
\end{tabular}
}
\caption{Early-training reward and gradient statistics. TaPR and outcome-only GRPO use the same Qwen3-8B initialization, APPS training data, optimizer, and batch configuration; only the reward definition differs. Zero-reward frequency is measured over the first 10 steps.}
\label{tab:reward-dynamics}
\end{table*}

\section{Ablation}
\label{sec:ablation}

To isolate the mechanism by which dense per-turn reward alters policy behavior, we compare TaPR against standard GRPO under identical architecture, data, optimizer, and KL-constraint settings.

\subsection{Reward Signal Properties}
\label{sec:reward-dynamics}

Table~\ref{tab:reward-dynamics} separates three stages of the early-training mechanism: whether a rollout produces a usable signal, the scale of that signal, and the resulting optimizer response. Because outcome-based GRPO yields zero reward whenever a candidate execution fails full test validation, its updates are substantially sparser than those of TaPR.

\begin{figure*}[t]
\centering
\includegraphics[width=0.72\textwidth]{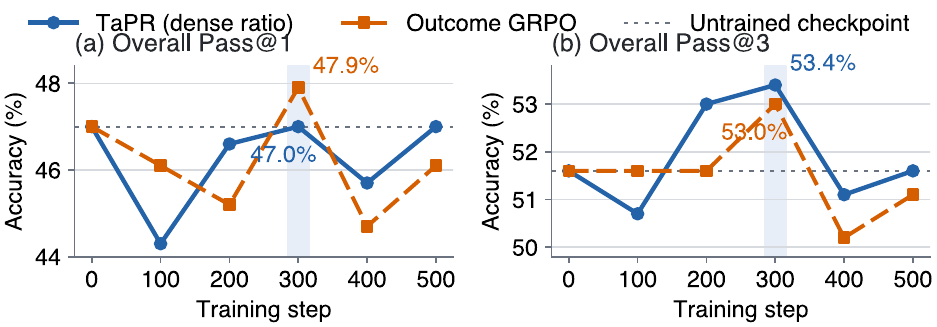}
\caption{Ablation overlay on Qwen3-8B: TaPR vs.\ GRPO trajectories on LCB-219 across training steps. TaPR reaches its pooled \passatk{3} peak by step 300; GRPO remains near baseline early and closes most of the pooled gap at step 300. Table~\ref{tab:passat3} reports the corresponding Hard-subset comparison.}
\label{fig:ablation}
\end{figure*}

Under matched initialization, TaPR produces a $13\times$ higher initial reward mean ($0.160$ vs.\ $0.0125$) and standard deviation ($0.320$ vs.\ $0.0250$) relative to GRPO. Early-stage policy gradient norms under TaPR are correspondingly $10\times$ larger ($0.030$ vs.\ $0.0030$). While GRPO encounters zero-reward signals in $40\%$ of the initial 10 steps, TaPR receives a nonzero signal at every one of these steps. These statistics establish denser early credit assignment; the checkpoint evaluations in Figure~\ref{fig:ablation} and Table~\ref{tab:passat3} determine whether that signal translates into evaluated behavior.

At the matched step-300 checkpoints, overall $\text{Pass}@1$ is $47.0\%$ for both the baseline and TaPR and $47.9\%$ for GRPO; all three score $7/61$ on the Hard subset. Performance divergence therefore emerges under multi-turn evaluation ($\text{Pass}@3$), where policies can leverage execution feedback.

\subsection{Multi-Turn Refinement Analysis ($\text{Pass}@3$)}
\label{sec:passat3}

Table~\ref{tab:passat3} compares the two reward definitions at the same step-300 training budget. All rows use Qwen3-8B and the same three-turn LCB-219 evaluation. The untrained row is the shared initialization; the two trained rows differ only in whether training uses final correctness or dense per-turn test-pass ratio.

\begin{table*}[t]
\centering
{\small
\begin{tabular}{lcccc}
\toprule
Training reward & \passatk{1} & \passatk{3} & Hard \passatk{1} & Hard \passatk{3} \\
\midrule
Untrained       & $47.0\%$ & $51.6\%$ & $11.5\%$ & $11.5\%$ \\
\rowcolor{resultgray}
Dense ratio (TaPR) & $47.0\%$ & $\mathbf{53.4\%}$ & $11.5\%$ & $\mathbf{16.4\%}$ \\
Final correctness (GRPO) & $47.9\%$ & $53.0\%$ & $11.5\%$ & $13.1\%$ \\
\bottomrule
\end{tabular}
}
\caption{Matched step-300 ablation on Qwen3-8B. Training uses the same APPS data and optimization configuration; evaluation uses the same 219 LCB problems and at most three feedback-conditioned attempts. Hard results are percentages over the 61-problem Hard subset.}
\label{tab:passat3}
\end{table*}

At step 300, outcome GRPO nearly matches TaPR on pooled $\text{Pass}@3$ ($53.0\%$ vs.\ $53.4\%$). On Hard problems, all three variants obtain $7/61$ with one attempt, whereas three-turn success is $7/61$, $8/61$, and $10/61$, respectively. The gap therefore appears in feedback-conditioned repair rather than first-attempt generation.

\paragraph{Joint interpretation.}
Tables~\ref{tab:reward-dynamics} and~\ref{tab:passat3}, together with Figures~\ref{fig:training-curve} and~\ref{fig:ablation}, separate signal availability, optimization trajectory, and evaluated behavior. The reward statistics show that TaPR exposes partial progress earlier; they do not alone establish a better final policy. The checkpoint curves reveal non-monotonicity and make step selection visible. At the matched step-300 budget, GRPO closes most of the pooled gap, while TaPR retains the higher Hard $\text{Pass}@3$ score with the same Hard $\text{Pass}@1$. The supported claim is therefore earlier and more visible refinement learning in this controlled setting, not universal superiority of dense reward.

\section{Discussion}
\label{sec:discussion}

\paragraph{Where the dense signal helps.}
TaPR is most informative when an initial program is incorrect but passes enough tests to expose partial progress. If a problem is already solved on the first attempt, later-turn reward cannot change $\text{Pass}@3$; if every candidate fails all tests, the pass ratio collapses to the same zero signal as outcome reward. The paired difficulty analysis is consistent with this middle regime. Medium problems contribute $23$ TaPR-only wins and $9$ baseline-only wins ($+2.99$\,pp), while Hard problems produce fewer discordant pairs but a directional $6{:}1$ win ratio. These strata are diagnostics rather than selection criteria: the reported 7B/8B slice is defined by model scale and refinement headroom, not by retaining favorable individual outcomes.

\paragraph{Limits of the observed effect.}
The two 14B models change by only $+0.91$ and $+0.46$\,pp, respectively. This may reflect less remaining refinement headroom, the capacity of the LoRA configuration, APPS-to-LCB distribution shift, or ordinary checkpoint variance; the present experiments do not distinguish among these explanations. The Qwen3-8B trajectory also returns to baseline by step 500, showing that a denser signal does not remove checkpoint sensitivity or overfitting. In addition, partial-test reward is only as reliable as the tests themselves. A policy can increase the pass ratio without learning a generally correct or secure solution when tests are incomplete.

\paragraph{Implications for evaluation.}
Multi-turn code studies should report first-attempt and feedback-conditioned outcomes together. $\text{Pass}@1$ alone misses the Hard-subset difference in Table~\ref{tab:passat3}, whereas pooled $\text{Pass}@3$ alone obscures that GRPO closes most of the aggregate gap by step 300. Paired wins and losses further distinguish a broadly distributed gain from a change driven by a small set of problems. Finally, public comparisons should preserve the benchmark release, split, sampling budget, and number of feedback turns. Table~\ref{tab:matrix} exposes these protocol fields rather than treating heterogeneous results as a controlled ranking.

\paragraph{Scope of the dissociation.}
Equal $\text{Pass}@1$ with different $\text{Pass}@3$ rules out an explanation based only on more first-attempt Hard successes. It does not prove that each later success is a local repair: feedback may trigger broad regeneration. The claim is behavioral---under the same protocol, TaPR converts more initially failed interactions into eventual successes. It also depends on test granularity. With one binary test, $P_t$ reduces to outcome reward; redundant tests can overstate progress on one behavior. The result therefore applies to the APPS and LCB test protocols studied here, not automatically to tasks without informative executable tests.

\section{Conclusion}
\label{sec:conclusion}
TaPR converts the change in hidden-test pass ratio after each turn into a dense reward, separating first-shot generation from feedback-conditioned repair. Across six paired LCB-219 evaluations, the selected checkpoints improve multi-turn $\text{Pass}@3$ by $+2.44$\,pp; the 7B/8B slice improves by $+3.31$\,pp, while the two 14B changes remain small.

The matched Qwen3-8B ablation localizes the difference. Baseline, outcome-only GRPO, and TaPR all score $7/61$ on Hard $\text{Pass}@1$, but their Hard $\text{Pass}@3$ scores are $7/61$, $8/61$, and $10/61$. GRPO nearly matches TaPR's pooled score by step 300, whereas TaPR supplies denser early credit and reaches the higher Hard-subset peak. The evidence supports faster refinement learning within this budget, not universal superiority of dense reward. Public comparisons in Table~\ref{tab:matrix} remain descriptive because their protocols differ; controlled comparisons should match checkpoint, benchmark, and interaction budget.

\section*{Ethical Statement}
TaPR may increase risks from overreliance on generated code. Training executes model-generated programs; runners should isolate processes, restrict filesystem and network access, and enforce resource limits. Hidden tests may miss unsafe behavior, so pass ratios do not establish correctness or security; human review remains necessary.

\bibliography{references}
\end{document}